\documentclass[conference]{IEEEtran}
\IEEEoverridecommandlockouts
\usepackage{cite}
\usepackage{amsmath,amssymb,amsfonts}
\usepackage{algorithmic}
\usepackage{graphicx}
\usepackage{textcomp}
\usepackage{xcolor}
\usepackage{url}
\graphicspath{{./}}

\definecolor{gg}{RGB}{0,100,0}

\def\BibTeX{{\rm B\kern-.05em{\sc i\kern-.025em b}\kern-.08em
    T\kern-.1667em\lower.7ex\hbox{E}\kern-.125emX}}
\begin{document}

\title{Texture Superpixel Clustering from Patch-based Nearest Neighbor Matching
}

\author{\IEEEauthorblockN{R{\'e}mi Giraud}
\IEEEauthorblockA{\textit{Signal and Image group} \\
\textit{Bordeaux INP, Univ. Bordeaux, CNRS, IMS,  UMR 5218}\\
F-33405 Talence, France \\\
remi.giraud@ims-bordeaux.fr}
\and
\IEEEauthorblockN{Yannick Berthoumieu}
\IEEEauthorblockA{\textit{Signal and Image group} \\
\textit{Bordeaux INP, Univ. Bordeaux, CNRS, IMS,  UMR 5218}\\
F-33405 Talence, France \\
yannick.berthoumieu@ims-bordeaux.fr}
}

\maketitle

\begin{abstract}
Superpixels are widely used in computer vision applications.
Nevertheless, decomposition methods may still fail to efficiently cluster image pixels 
according to their local texture.
In this paper, 
we propose a new Nearest Neighbor-based Superpixel Clustering (NNSC) method to generate 
texture-aware superpixels in a limited computational time compared to previous approaches.
We introduce a new clustering framework using patch-based nearest neighbor matching, while
most existing methods are based on a pixel-wise K-means clustering.
Therefore, we directly group pixels in the patch space enabling to capture texture information.
We demonstrate the efficiency of our method with favorable comparison
in terms of segmentation performances on both standard color and texture datasets.
We also show the computational efficiency of NNSC compared to recent texture-aware superpixel methods.
\end{abstract}


\section{Introduction}

The constant increase of image data may highly impact the computational cost 
of computer vision pipelines. 
Among the approaches used to reduce the processing load, 
dimension reduction and multi-resolution methods have been widely used over the past years.
In this context, decompositions into superpixels appear to be very interesting,
since the created regions tend to respect the boundaries of the image objects.
The relations between these irregular regions at different resolution levels can still be inferred \cite{nakamura2017hierarchical}, and 
superpixel neighborhoods can be used as for standard  regular patch-wise and  multi-resolution processing
\cite{giraud2017_spm}.
Therefore, many superpixel-based methods have been proposed in the literature, for different image processing and analysis applications,
\emph{e.g.},
semantic segmentation \cite{mostajabi2015feedforward},
tracking with optical flow \cite{menze2015object}, %
depth estimation \cite{zitnick2007stereo}, and 
color \cite{rabin2014non} or style transfer \cite{liu2018photo}.

Since their introduction and popularization with \cite{achanta2012},
the majority of superpixel methods decompose the image into regions approximately containing the same number of pixels with homogeneous colors. 
To compute this clustering, most methods such as
\cite{liu2011,vandenbergh2012,achanta2012,yao2015,zhang2017ssgd,chen2017,achanta2017superpixels,giraud2018_scalp},
consider a trade-off distance between spatial and color spaces at the pixel scale.
The spatial distance enables to provide a relatively regular decomposition of the image domain, 
while the color distance associates pixels to a superpixel with the same average color.
Most state-of-the-art methods only use the pixel spatial and color features in their clustering model, since information at the pixel scale may be sufficient to detect the object boundaries in a natural color image. 
Consequently, recent works such as \cite{stutz2018,giraud2019_tasp}
have highlighted the non robustness of pixel-wise state-of-the-art methods to 
noise or texture for instance.
All methods relying on pixel-wise information may indeed highly fail at grouping textures and may provide very inconsistent decompositions \cite{giraud2019_tasp}.
Even recent methods using advanced feature spaces \cite{liu2016manifold,chen2017} or
additional information such as features on the path to the superpixel barycenter  \cite{buyssens2014,zhang2017ssgd,giraud2018_scalp}
do not explicitly capture texture patterns and fail to detect texture changes.

In the recent method TASP \cite{giraud2019_tasp}, a straightforward extension of the SLIC framework \cite{achanta2012} is proposed to compute texture-aware superpixels.
Patch comparisons are performed within the superpixel to provide a texture term in the clustering model.
Nevertheless, such method presents an important computational complexity.
The SLIC framework \cite{achanta2012} begin based on a K-means clustering, each superpixel iteratively computes its distance to all pixels in a restricted area.
In TASP \cite{giraud2019_tasp}, the texture term must be computed for each superpixel at each iteration, by a nearest neighbor search performed for all pixels in this area, leading to an important computational burden, \emph{i.e.}, more than $60$s for images of $321{\times}481$ pixels.
Hence, it appears necessary to propose a more efficient approach in terms of complexity, also able to accurately cluster textures.

\vspace{0.05cm}

\subsubsection*{Contributions}
In this paper, 
we propose a new Nearest Neighbor-based Superpixel Clustering (NNSC) method to generate accurate and texture-aware superpixels. 
We introduce a new clustering framework using patch-based nearest neighbor matching, while
most existing methods are based on a K-means clustering.
Hence, we directly group pixels in the patch space while previous methods such as \cite{giraud2019_tasp}, 
 combine both approaches at the expense of an important computational load.
We also propose a new method to merge several decomposition estimations obtained
from different nearest neighbor searches.

In the following, we first show
the interest of considering patches for texture clustering, 
and the limitations of their use in a K-means-based clustering algorithm \cite{giraud2019_tasp}.
Then, we present our new decomposition method relying on a patch-based nearest neighbor clustering, 
with much lower complexity.
Finally, we study our method parameters and compare its segmentation performances 
to the ones of the state-of-the-art methods on both natural color and texture datasets.

\section{Texture Superpixels using Patches}

In this section, we first demonstrate the ability of patches
to easily  cluster image textures.
Then, we present the standard pixel-wise K-means-based clustering algorithm \cite{achanta2012}
and the limitations of its extension using patches to generate texture-aware superpixels in \cite{giraud2019_tasp}.

\subsection{Texture Clustering using Patches}

\begin{figure}
\begin{center}
\begin{tabular}{@{\hspace{0mm}}c@{\hspace{5mm}}c@{\hspace{0mm}}}
  \includegraphics[width=0.2425\textwidth,height=0.18\textwidth]{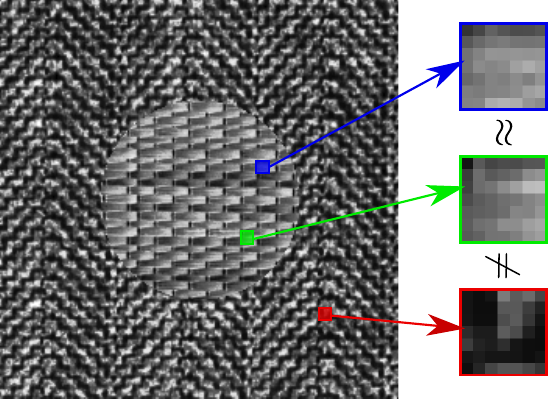} & 
  \includegraphics[width=0.2175\textwidth,height=0.18\textwidth]{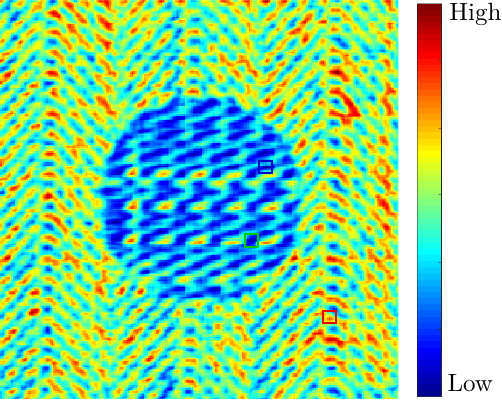} \\[-0.5ex]
\end{tabular}
\end{center}
 \caption{Interest of patches for texture clustering. 
 Selection of a reference patch of size $7{\times}7$ pixels in blue (left).
 A similar (green) and distant (red) patches are represented.  
 Map of $l$-2 distance \eqref{patch} between the reference blue patch and all image patches  (right), demonstrating the patch ability to capture regions with the same texture. 
 } \vspace{-0.15cm}
 \label{fig:patch}
\end{figure}

Patches enable to capture the neighborhood of each image pixel.
Non-local patch-based approaches, that have first become popular for texture synthesis \cite{efros99} and denoising applications \cite{buades2005},
use this structure to find similar patterns in the same or other images.
The distance between fixed size patches enables to reflect both the
similarity in terms of intensity and texture patterns.
This distance $d_P$ between two patches $P(p_i)$ and $P(p_k)$  describing the neighborhood of two pixels $p_i$ and $p_k$, is generally computed with a $l$-2 norm such that:  

{\small
\begin{equation}
 d_P(p_i,p_k)=\frac{1}{n}{\|P(p_i)-P(p_k)\|}_2 , \label{patch}
\end{equation}
}%
with $n$ the patch size. 
In Figure \ref{fig:patch}, we illustrate the ability of patches to cluster image textures
by computing distance \eqref{patch} between a reference patch and all other image patches.

In the context of texture-aware superpixel clustering,  
the texture homogeneity between a pixel and a superpixel is not easy to measure since a pixel neighborhood must be compared to a superpixel having a variable size.
Texture classification approaches could necessitate prior information on the image type, or additional parameter settings to be consistent with the pixel-wise color information that must also be taken into account in the clustering model \cite{giraud2019_tasp}.
Moreover, such approaches can be computationally costly.
Therefore, using patches appears to be an interesting solution but requires a selection strategy to determine which patches to compare.

\subsection{Texture Superpixels from K-means-based Framework}

The recent TASP method \cite{giraud2019_tasp} proposes to
generate texture-aware superpixels using the  K-means-based framework of \cite{achanta2012},
which is 
very popular due to its simplicity of use and understanding.
The image is first split into regular blocks of size $s{\times}s$, 
depending of the input number of desired superpixels.
Superpixels are then sequentially processed, and try to gather neighboring pixels in a restricted area of size $(2s+1){\times}(2s+1)$.
The clustering distance between a pixel and a superpixel
is composed of a spatial and color distance.
The pixel features are compared to the average features over all pixels in the superpixel.
At the end of each iteration, pixels are associated to the superpixel providing the lowest distance.

The TASP method \cite{giraud2019_tasp} adds a texture homogeneity term to the  distance of \cite{achanta2012}.
It uses fixed size patches as descriptors to easily capture texture patterns while staying in the same feature space as the color distance between pixels and superpixels.
Figure \ref{fig:patch} shows that patch distances may be high even within the same texture area.
Therefore, comparing a pixel neighborhood described by a patch to a reference one, for instance at the superpixel barycenter would not guarantee a relevant texture measure.
Hence, \cite{giraud2019_tasp} performs a patch-based nearest neighbor (NN) search to find similar patches in the superpixel.
Similar patches then implies texture homogeneity, and favor the association of the pixel to the superpixel.

\vspace{0.1cm}

\subsubsection*{Limitations} 
With such approach, the NN search must be performed for all pixels in the $(2s+1){\times}(2s+1)$ pixels  area  for each superpixel at each iteration, leading to 
 overlapping pixels and repetition of the NN matching process.
In the K-means-based clustering, a pixel is indeed  approximately considered by $4$ superpixels at each iteration.
Therefore, TASP complexity depends on the number of image pixels $|I|$, number of K-means iterations $N_{K}$, and number of NN search iterations $N$ such that $C_{\text{TASP}} = \mathcal{O}(|I|{\times}4{\times}N_{K}{\times}N)$.

These limitations motivate the introduction of our new clustering framework, significantly  reducing this complexity while preserving the ability to generate texture-aware superpixels.

\section{Texture Superpixels from Patch-based Nearest Neighbor Matching}

In this section, we first introduce our new clustering framework directly based on NN matching.
Then, we present in detail the algorithm used to perform the search of similar patches.
Finally, we propose a method to merge several superpixel decompositions  obtained from different NN matching.

\subsection{Nearest Neighbor Superpixel Clustering Framework}

 \subsubsection{Clustering Algorithm}
 The proposed NNSC method directly clusters pixels using a patch-based NN matching process, that we prove to be necessary to provide texture-aware superpixels. 
 The search is sequentially performed for all image pixels, to iteratively refine the initial superpixel grid decomposition.
 Therefore, it differs from the standard K-means-based framework that sequentially processes superpixels, leading the same pixel to be considered several times at the same iteration.

\begin{figure*}
\begin{center}
  \includegraphics[width=0.925\textwidth,height=0.315\textwidth]{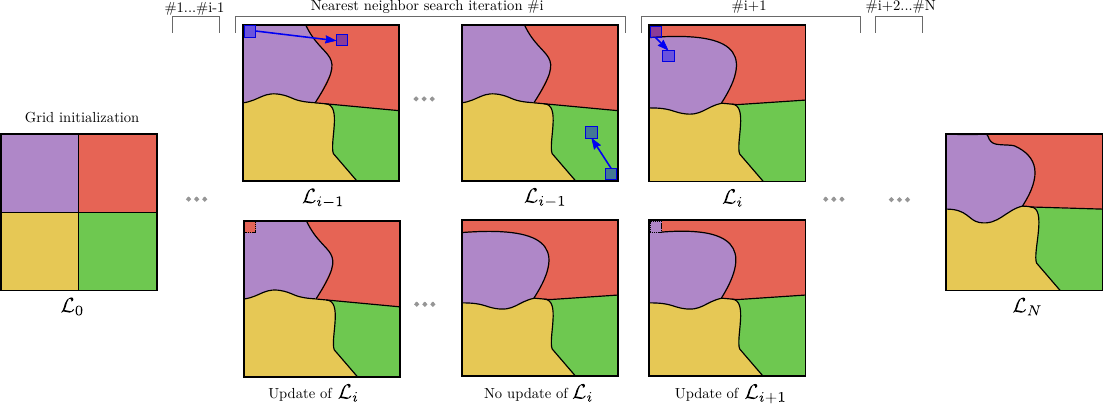}
\end{center}\vspace{-0.375cm}
 \caption{
 Illustration of the NNSC clustering framework based on NN matching for a number of $4$ superpixels.
 Superpixels are first decomposed into a regular grid.
 At each iteration $\#i$, pixels in blue are sequentially considered, and a correspondence is found within the image (lines), using patch-based distances \eqref{psc_dist}.
 If a pixel gets associated to a different label, the $i$-th label map $\mathcal{L}_i$ is updated.
 After $N$ iterations, the final label map $\mathcal{L}_N$ is obtained.
 } \vspace{-0.2cm}
\label{fig:psc}
\end{figure*}

 The NNSC decomposition process to obtain a label map $\mathcal{L}$ is illustrated in Figure \ref{fig:psc}.
 At a given iteration, the label of the superpixel containing the patch correspondence is assigned to the considered pixel position for next iteration.
 The complexity of NNSC with its clustering framework directly based on NN matching reduces to
 $C_{\text{NNSC}} = \mathcal{O}(|I|{\times}N)$.
Note that the search for similar patches can be performed by any NN method, and we present the proposed search strategy in section \ref{sec:pm}.

 \subsubsection{Patch-based Clustering Distance}

 To capture both the similarity in terms of intensity and texture patterns,
 patch intensities in the feature space (\emph{e.g.}, colors in CIELab color space) are considered in the patch-based distance $d_P$
computed between a pixel $p_i$ of patch $P(p_i)$, and a patch $P(p_k)$ at position $p_k\in S_k$ such that: 
 \vspace{-0.3cm}
 
{\small
\begin{equation}
d_P(p_i,p_k) = \frac{1}{n}{\|P(p_i)-P(p_k)\|}_2  + \frac{m_{k}^2}{s^2}\hspace{0.05cm} \Gamma \big(p_k,X_{S_k}\big) ,  \label{unicity}
\end{equation}
}%
\noindent with
$m_k$ the regularity parameter, automatically 
set for each superpixel $S_k$ \cite{giraud2019_tasp}, and $\Gamma$, a spatial weighting function defined such that  $\Gamma(p_k,X_{S_k}) = 2s^2(1-\exp{(-{\|{p_k}-X_{S_k}\|}_2^2/s^2)})$, 
 favoring the search near to the superpixel barycenter $X_k$, preventing a superpixel $S_k$ to cluster different textures.

Finally, the global patch-based clustering distance $D$ considers the patch distance term $d_P$ \eqref{unicity}, but also the standard color $d_c$ and spatial $d_s$ distances at the pixel scale \cite{achanta2012}.
These terms respectively enable to adapt superpixel borders to object contours and to ensure the shape regularity of superpixels.
Hence, patch correspondences are computed according to: \vspace{-0.3cm}

{\small
\begin{equation}
D(p_i,p_k) =  d_P(p_i,p_k) + d_c(p_i,S_k) + d_s(p_i,S_k)\frac{m_k^2}{s^2} . \label{psc_dist}
\end{equation}
}

\vspace{-0.45cm}

\subsection{\label{sec:pm}Nearest Neighbor Search using PatchMatch}

Since computing exact NN would be too costly,
we choose to use the approximate NN search algorithm PatchMatch (PM) \cite{barnes2009}.
PM was initially proposed to provide for each patch of an image $A$, a correspondence in an image $B$. 
The algorithm starts from random correspondences and iteratively refines the patch associations using fast propagation of good matches from adjacent neighbors, and random tests.
We adapt this algorithm to our context, \emph{i.e.}, finding similar patches within the same image and into a restricted area around each patch.

\begin{figure} 
{\footnotesize 
\begin{center}
 \begin{tabular}{@{\hspace{0mm}}c@{\hspace{1.5mm}}|c@{\hspace{3mm}}c@{\hspace{0mm}}}
 & \multicolumn{2}{c}{Iteration \#1}\\
   \includegraphics[width=0.15\textwidth]{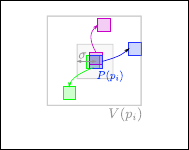} & 
   \includegraphics[width=0.15\textwidth]{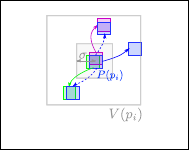} & 
   \includegraphics[width=0.15\textwidth]{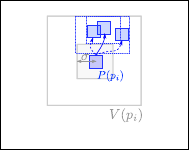} \\
   (a) Initialization & (b) Propagation & (c) Random search\\[-1ex]
 \end{tabular}
\end{center}
}
 \caption{Illustration of the PM algorithm adapted to the search of approximate NN within the same image.
 The processing is described for a blue patch.
 (a) Random initialization for the blue patch and two of its adjacent neighbors within a constrained window and outside a $\sigma$-neighborhood.
 (b) The propagation tests the shifted correspondences of recently processed adjacent patches (dotted lines). 
(c) The random search performs random tests within an iteratively reducing window around the current best match.
} \vspace{-0.3cm}
\label{fig:pm}
\end{figure}

First, to ensure the regularity of the decomposition, we limit the search to a $(2s+1){\times}(2s+1)$ pixels area around the pixel position.
For a pixel $p_i$, this area is denoted $V(p_i)$ in Figure \ref{fig:pm}, which illustrates the algorithm steps for a given patch $P(p_i)$.
Naturally, to avoid to match the same patch $P(p_i)$, 
a $\sigma$-neighborhood is defined where to prevent the selection of patches.
Random associations are first computed in these restricted areas (Figure \ref{fig:pm}(a))
after the grid initialization.
Then, the propagation step considers the correspondences of the recently processed adjacent patches to lead $P(p_i)$ to new potential correspondences (Figure \ref{fig:pm}(b)).
Finally, random selections are performed in areas of reducing size around the best current correspondence (Figure \ref{fig:pm}(c)).
This adaptation of PM finds similar patches in the same image while spatially constraining the search area to ensure superpixel regularity.

\subsection{Aggregation of Multiple Clustering Estimation}

PM being partly random, several clustering estimations can be computed, 
and averaged to improve the performances, as in \cite{giraud2016_opal}.
These independent estimations can be easily launched in parallel using multi-threading implementation.
Variations between estimations being reduced, 
the aggregation of $M$ multiple label maps $\mathcal{L}^i_N$ can be performed as follows:  \vspace{-0.1cm}

{\small
\begin{equation}
 \mathcal{L}_{{final}}(p_i) = \underset{l\in \{labels\}}{\text{argmax}} \text{ } \sum_{i=1}^{M} \delta_{\mathcal{L}_N^i(p_i),l} \text{ }, \label{aggreg}
\end{equation}
}
where $\delta_{i,j}$ equals $1$ when $i=j$, and $0$ otherwise.

Finally, 
as in \cite{achanta2012}, 
a post-processing step ensuring superpixel connectivity is performed on the final label map  $\mathcal{L}_{{final}}$.

\section{Evaluation of Performances}

\subsection{Validation Framework}

\subsubsection{Dataset} 
To evaluate the segmentation performances, we
consider a standard composite texture image (CTI) dataset \cite{randen1999filtering}, which 
is composed of $10$ grayscale images containing up to $16$ different textures\footnote{Dataset available at: \url{http://rgiraud.vvv.enseirb-matmeca.fr/nnsc/}}. 
High performances on these images demonstrate the ability to detect texture changes.
We also report the performances for the standard natural color Berkeley Segmentation Dataset (BSD) \cite{martin2001},
which contains $200$ test images of size $321{\times}481$ pixels.

\subsubsection{Compared methods} 
NNSC performances are compared to the ones of the recent state-of-the-art methods
 SLIC \cite{achanta2012},
 ERGC \cite{buyssens2014},
 ETPS \cite{yao2015},
 LSC \cite{chen2017},
 SNIC \cite{achanta2017superpixels},
 SCALP \cite{giraud2018_scalp},
 and TASP \cite{giraud2019_tasp},
 used with parameters recommended by the authors.
Performances are measured with the standard Achievable Segmentation Accuracy (ASA) \cite{liu2011}
that evaluates the accuracy of superpixels according to a ground truth segmentation.

\subsubsection{Parameters}
In NNSC default settings, patches of size $n=7{\times}7$ pixels are selected outside a 
$\sigma=3$ neighborhood.
$M=4$  label maps are aggregated, and the number of iterations is set to $8$.
Features and parameters in \eqref{psc_dist} are computed as in \cite{giraud2019_tasp}.
These parameters are empirically set, and results in section \ref{sec:comp_soa} are obtained using the same settings.
Finally, note that the random sequence of PM is controlled to provide the same decomposition for the same image and parameters.


\subsection{Influence of Parameters}

 \subsubsection{Patch Size}

\begin{figure} 
{\footnotesize 
\begin{center}
 \begin{tabular}{@{\hspace{0mm}}c@{\hspace{2mm}}c@{\hspace{0mm}}}
   \includegraphics[width=0.235\textwidth,height=0.195\textwidth]{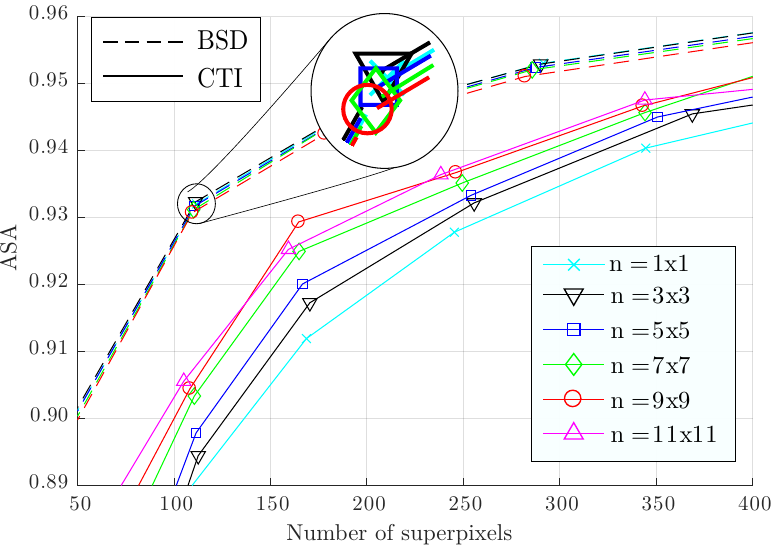} & 
      \includegraphics[width=0.235\textwidth,height=0.195\textwidth]{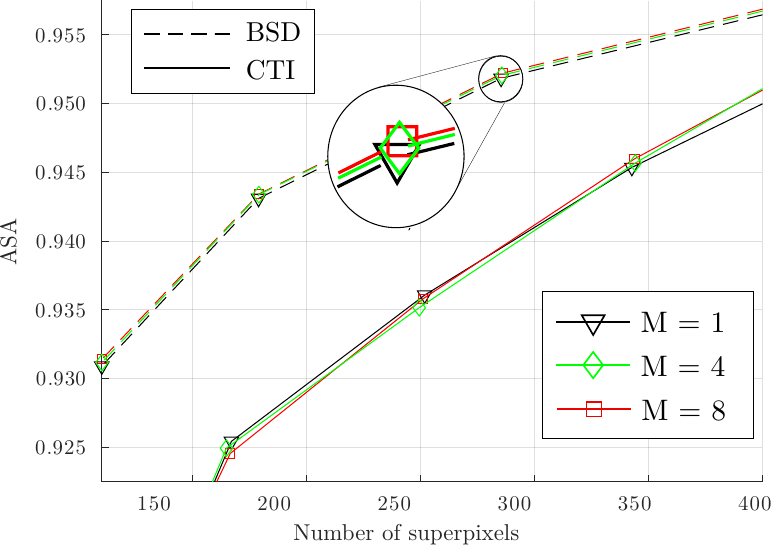} \\
   (a)&(b)\\[-1.25ex]
 \end{tabular}
\end{center}
}
 \caption{Influence of (a) the patch size \eqref{unicity} and
 (b) the number of aggregated label maps estimated from different NN searches \eqref{aggreg} on ASA.
}\vspace{-0.2cm}
\label{fig:param}
\end{figure}

The influence of the patch size \eqref{unicity} on the performances
is shown in Figure \ref{fig:param}(a).
On the CTI dataset, large patches enable to efficiently capture textures, while
on the BSD dataset patches larger than $3{\times}3$ do not provide more information, object contours being mainly detected by color changes.
In NNSC default settings, a patch size of $7{\times}7$ is chosen as a good trade-off between accuracy and computational time.
Nevertheless, parameters could be manually optimized.

 \subsubsection{Number of Clustering Estimations}
 The influence of the number $M$ of aggregated label maps from different NN searches \eqref{aggreg}  is shown in Figure \ref{fig:param}(b).
Aggregating several estimates enables to improve the segmentation performances by smoothing the decision at superpixel boundaries. 
A reduce number of $M=4$ label maps is chosen in the following.

\subsection{\label{sec:comp_soa}Comparison to the State-of-the-Art Methods}

\subsubsection{Segmentation Performances}

\begin{figure}[t]
{\footnotesize 
\begin{center}
 \begin{tabular}{@{\hspace{0mm}}c@{\hspace{2mm}}c@{\hspace{0mm}}}
   \includegraphics[width=0.235\textwidth,height=0.195\textwidth]{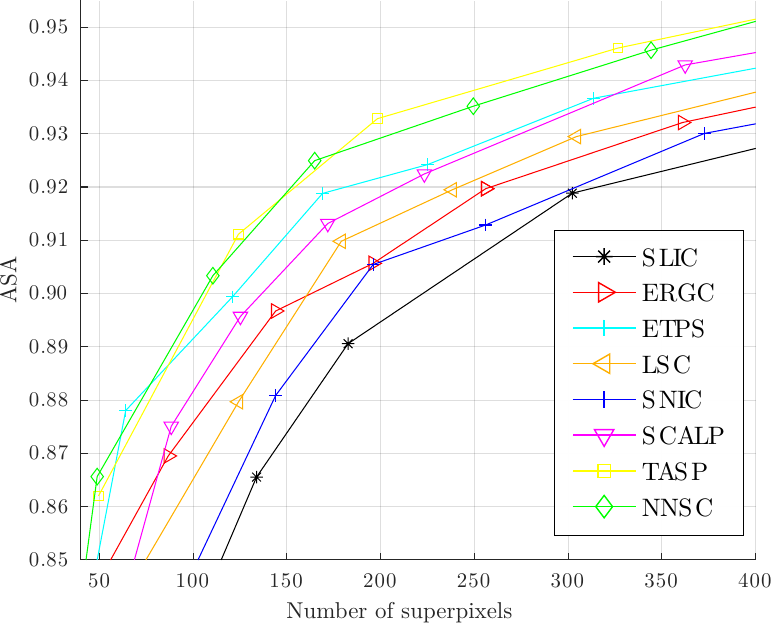}&
      \includegraphics[width=0.235\textwidth,height=0.195\textwidth]{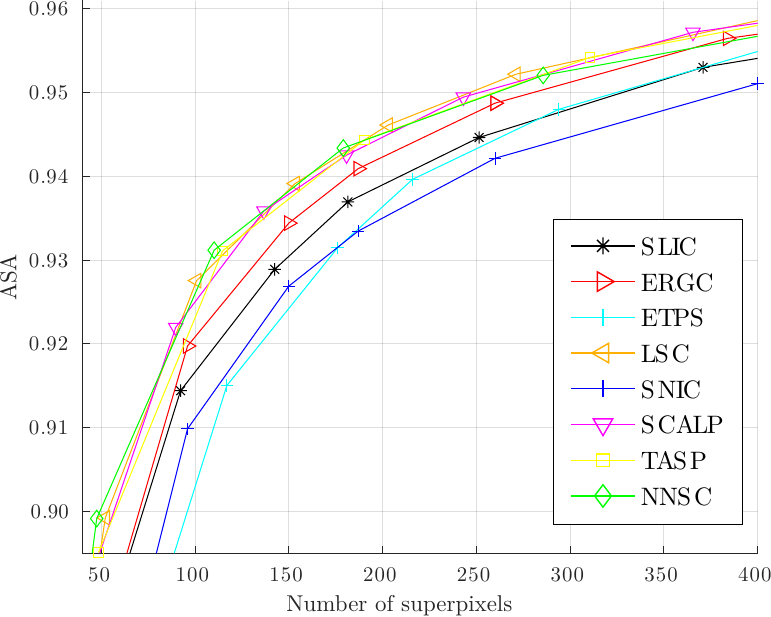}\\
         (a)&(b)\\[-1.25ex]
 \end{tabular}
\end{center}
}
 \caption{Comparison of NNSC segmentation performances measured with ASA, to the ones of the state-of-the-art methods on texture CTI (a) and natural color BSD (b) datasets.
} \vspace{-0.2cm}
\label{fig:asa_soa}
\end{figure}

Performances are reported for several superpixel scales in Figure \ref{fig:asa_soa}.
 NNSC obtains performances similar to TASP \cite{giraud2019_tasp} on the CTI dataset  Figure \ref{fig:asa_soa}(a), showing its capacity to produce texture-aware superpixels.
 while performing as well or better than the best compared methods on the BSD  Figure \ref{fig:asa_soa}(b).
Note that these results are obtained using the same parameters.

\begin{figure*}
\renewcommand{\arraystretch}{0.95}
\newcommand{\ww}{0.155\textwidth}
\newcommand{\hh}{0.105\textwidth}
\newcommand{\hhh}{0.1475\textwidth}
\newcommand{\hhhh}{0.11\textwidth}
\centering
{\color{black}
{\footnotesize
 \begin{tabular}{@{\hspace{1mm}}c@{\hspace{2mm}}c@{\hspace{2mm}}c@{\hspace{2mm}}c@{\hspace{2mm}}c@{\hspace{2mm}}c@{\hspace{0mm}}}
   \includegraphics[width=\ww,height=\hhh]{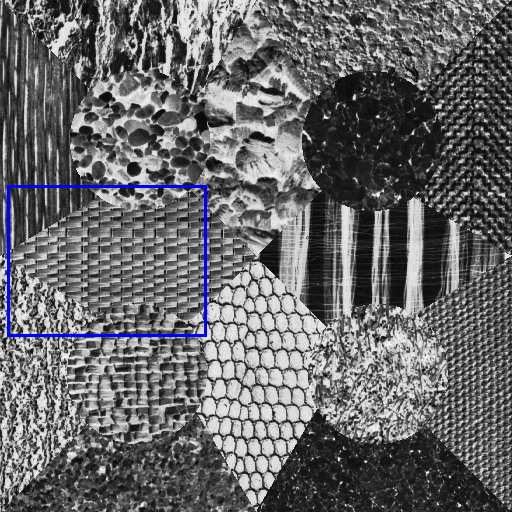}& 
  \includegraphics[width=\ww,height=\hhh]{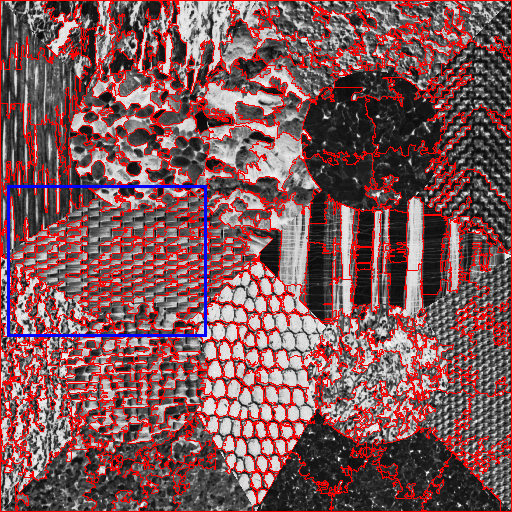}&
  \includegraphics[width=\ww,height=\hhh]{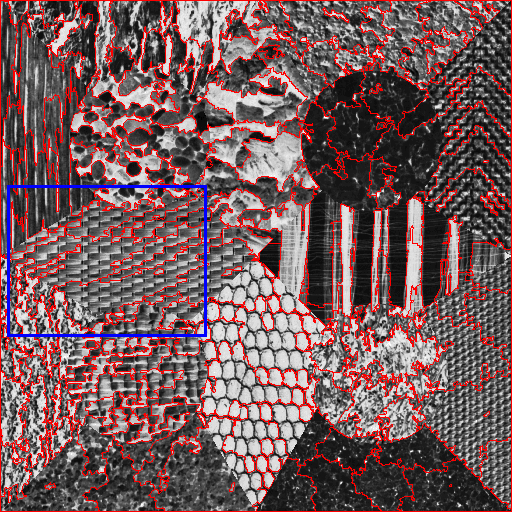}&
  \includegraphics[width=\ww,height=\hhh]{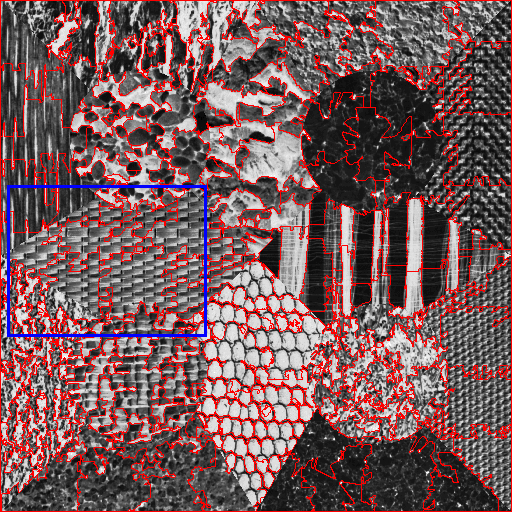}& 
  \includegraphics[width=\ww,height=\hhh]{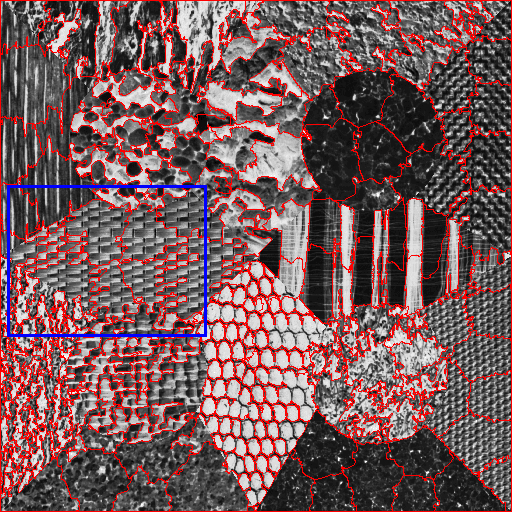}& 
  \includegraphics[width=\ww,height=\hhh]{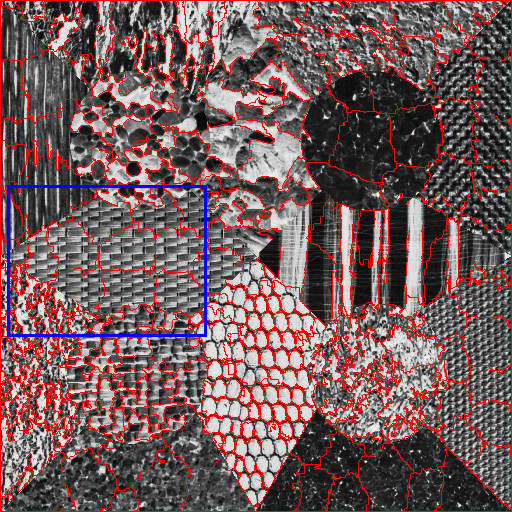}\\ 
   \includegraphics[width=\ww,height=\hhhh]{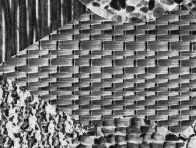}& 
  \includegraphics[width=\ww,height=\hhhh]{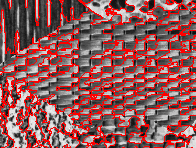}&
  \includegraphics[width=\ww,height=\hhhh]{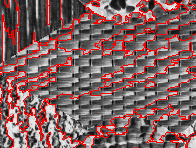}&
  \includegraphics[width=\ww,height=\hhhh]{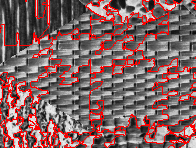}& 
  \includegraphics[width=\ww,height=\hhhh]{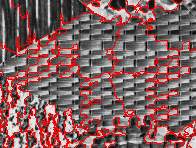}& 
  \includegraphics[width=\ww,height=\hhhh]{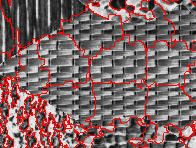}\\[0.75ex] 
   \includegraphics[width=\ww,height=\hh]{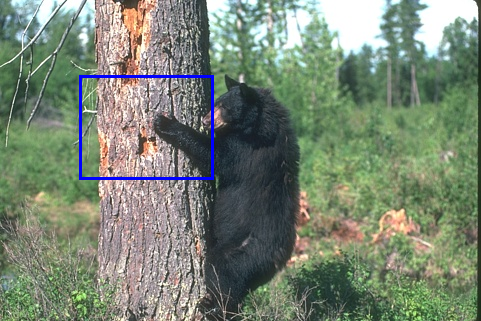}& 
  \includegraphics[width=\ww,height=\hh]{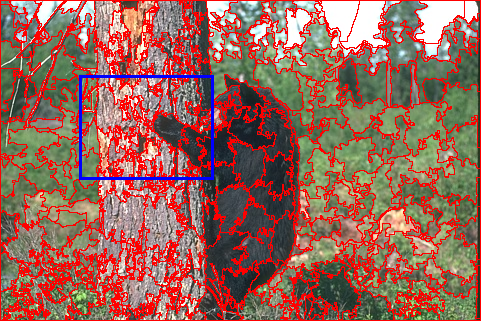}&
  \includegraphics[width=\ww,height=\hh]{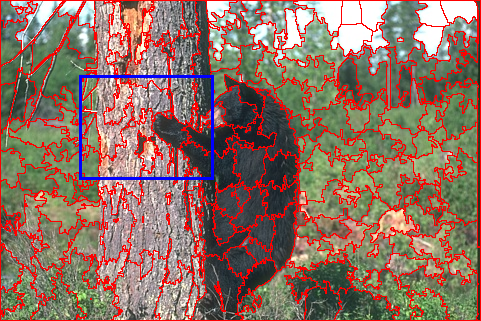}&
  \includegraphics[width=\ww,height=\hh]{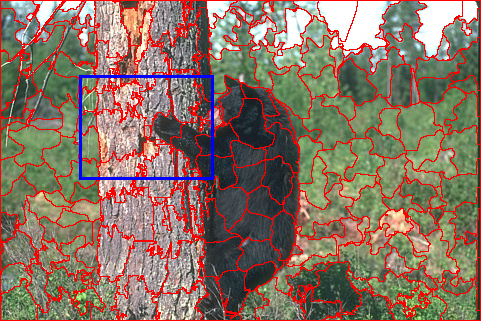}& 
  \includegraphics[width=\ww,height=\hh]{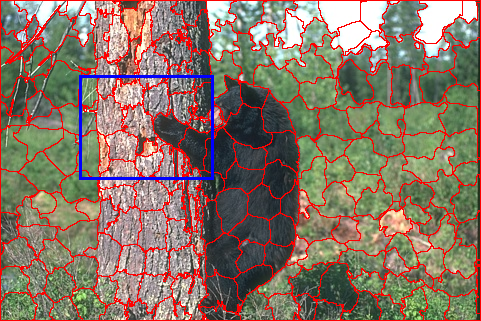}& 
  \includegraphics[width=\ww,height=\hh]{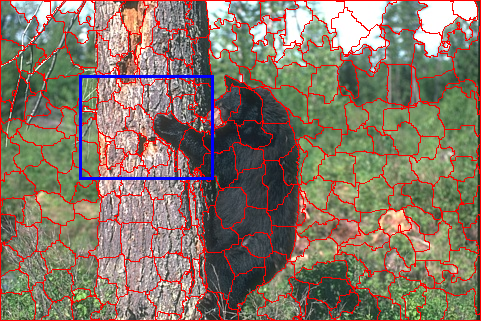}\\ 
   \includegraphics[width=\ww,height=\hh]{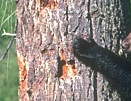}& 
  \includegraphics[width=\ww,height=\hh]{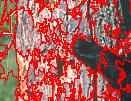}&
  \includegraphics[width=\ww,height=\hh]{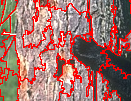}&
  \includegraphics[width=\ww,height=\hh]{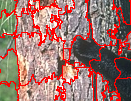}& 
  \includegraphics[width=\ww,height=\hh]{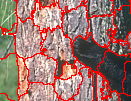}& 
  \includegraphics[width=\ww,height=\hh]{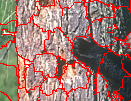}\\ 
  Initial image &LSC \cite{chen2017}  & SNIC \cite{achanta2017superpixels}  & SCALP \cite{giraud2018_scalp}&TASP  \cite{giraud2019_tasp} & NNSC \\[-0.75ex]
 \end{tabular}
 }
 }
\caption{
Visual comparison of the proposed NNSC method to the most recent state-of-the-art approaches on a CTI (top) and BSD example (bottom) for $200$ superpixels. 
NNSC provides as accurate or better decompositions with much less fuzzy superpixels on both texture composite and natural color images.  
}  \vspace{-0.25cm} 
\label{fig:comp_soa} 
\end{figure*}

 NNSC is also visually compared to the most recent state-of-the-art approaches in Figure \ref{fig:comp_soa}.
On the natural color image, NNSC provides relevant superpixels that accurately detect structures, \emph{e.g.}, the tree or the bear's arm.
On the complex composite texture image, NNSC provides more accurate segmentation, with much less fuzzy superpixel shapes.

\subsubsection{Computational Complexity}

NNSC presents a significantly reduced complexity
compared to the TASP texture-aware superpixel approach \cite{giraud2019_tasp}, whose
complexity depends on
the number of image pixels $|I|$, number of K-means iterations $N_K$, 
and number of NN search iterations $N$ such that,
$C_{\text{TASP}} = \mathcal{O}(|I|{\times}4{\times}N_{K}{\times}N)$, 
while $C_{\text{NNSC}} = \mathcal{O}(|I|{\times}N)$, 
since it is directly based on a NN clustering framework.

NNSC takes around $2$s in its default settings, while
TASP requires in average $60$s to decompose a BSD image of $321{\times}481$ pixels on a linux computer with 4 cores at 1.90GHz and 16GB of RAM.
With costly patch-based distances to handle textures, and without advanced code optimizations, 
NNSC achieves computational times similar to the ones of accurate methods such as \cite{liu2011,giraud2018_scalp}.
Finally, NNSC could reach real-time performances since 
several works have proposed such PM implementations using GPU architectures 
\cite{nover2018espresso}.

\section{Conclusion}

In this work, we propose a new superpixel method considering information at the patch scale
to cluster pixels having similar local texture properties.
The proposed approach iteratively clusters pixels using
a locally constrained patch-based nearest neighbor matching.
This way, it significantly reduces the complexity of existing texture-aware 
approaches, while preserving the accuracy of segmentation.
Future works will focus on the extension of the proposed method 
to 3D supervoxel decomposition,
with real-time processing, for
applications such as object tracking on video.

\end{document}